# General Backpropagation Algorithm for Training Second-order Neural Networks

Fenglei Fan, Wenxiang Cong, Ge Wang
Biomedical Imaging Center, BME/CBIS
Rensselaer Polytechnic Institute, Troy, New York, USA

*Abstract*— **The artificial neural network is a popular framework in machine learning. To empower individual neurons, we recently suggested that the current type of neurons could be upgraded to $2^{nd}$ order counterparts, in which the linear operation between inputs to a neuron and the associated weights is replaced with a nonlinear quadratic operation. A single $2^{nd}$–order neurons already has a strong nonlinear modeling ability, such as implementing basic fuzzy logic operations. In this paper, we develop a general backpropagation (BP) algorithm to train the network consisting of $2^{nd}$-order neurons. The numerical studies are performed to verify of the generalized BP algorithm.**

*Index Terms* — **Machine learning, artificial neural network (ANN), $2^{nd}$ order neurons, backpropagation (BP).**

## I. Introduction

In machine learning, artificial neural networks (ANNs) especially deep neural networks (CNNs) have been very successful in different applications including multiple areas in biomedical engineering such as medical imaging [1-3]. Usually, a neural network has several layers of neurons. For each neuron, the inner product between an input vector and its weighting vector is nonlinearly processed by an activation function such as Sigmoid or ReLU [4-6]. The model of the current neuron, by definition, cannot perform nonlinear classification unless a number of neurons are connected to form a network.

To enrich the armory of neural networks at the cellular level, recently we upgraded inner-product-based ($1^{st}$-order) neurons to quadratic-function-based ($2^{nd}$-order) neurons by replacing the inner product with a quadratic function of the input vector. In our feasibility study [7], a single $2^{nd}$-order neuron performed effectively in linearly inseparable classification tasks; for example, basic fuzzy logic operations. That is to say, the $2^{nd}$-order neuron has a representation capability superior to that of the $1^{st}$–order neuron.

While either $1^{st}$-order or $2^{nd}$-order neurons can be interconnected to approximate any functions [8], we hypothesize that to perform a complicated task of machine learning the network made of $2^{nd}$-order neurons could be significantly simpler that that with $1^{st}$-order neurons. To test this hypothesis, the training algorithm, known as backpropagation (BP), for traditional $1^{st}$-order neural networks needs to be extended to handle $2^{nd}$-order neural networks, although the principle of BP is essentially the same.

This work is partially supported by the Clark & Crossan Endowment Fund at Rensselaer Polytechnic Institute, Troy, New York, USA.

As we know, the key idea behind the training of ANNs is to treat it as an optimization problem [9]. The network learns the features when the optimization is accomplished with respect to the training dataset. In the optimization process, the propagation and backpropagation steps are repeated for gradual refinement until convergence. While the propagation produces an error of a loss function, the backpropagation calculates the gradients of the loss function with respect to weights, and minimizes the loss function by adjusting the weights. In doing so, the forward and backward steps are performed layer-wise according to the chain rule for differentiation.

In the next section, we formulate a general backpropagation algorithm for training of $2^{nd}$-order networks. In the third section, we present several numerical examples to show the feasibility and merits of $2^{nd}$–order networks relative to that with $1^{st}$–order neurons. Finally, we discuss relevant issues and conclude the paper.

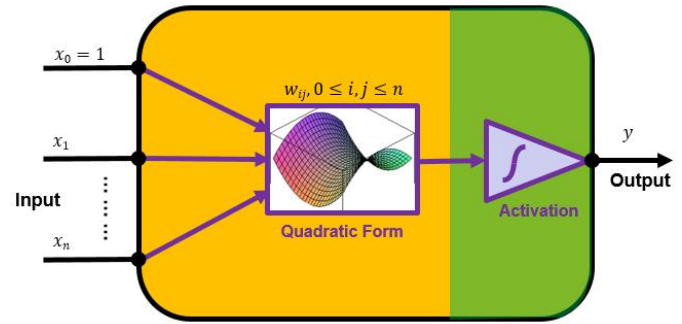

Fig. 1. Structure of our proposed $2^{nd}$-order neuron consisting of a quadratic function and an activation function, both of which are nonlinear.

## II. Backpropagation through the $2^{ND}$-order network

The structure of the $2^{nd}$-order neuron is shown in Fig. 1, where $w_{0\_r} \coloneqq b_1$, $w_{0\_g} \coloneqq b_2$ and $x_0 = 1$. The quadratic transform of the input vector is defined as follows [7]:

$$h(\vec{x}) = (\sum_{i=1}^{n} w_{i\_r} x_i + b_r)(\sum_{i=1}^{n} w_{i\_g} x_i + b_g) + \sum_{i=1}^{n} w_{i\_b} x_i^2 + c. \quad (1)$$

Then, $h(\vec{x})$ will be further processed by a nonlinear activation function, such as a Sigmoid or ReLU function. That is, in a $2^{nd}$-order neuron, the input vector is quadratically, instead of linearly, mapped before being fed into the excitation function. The $2^{nd}$-order neuron is e a natural extension of the $1^{st}$-order neuron. It is expected that the $2^{nd}$-order networks should be more efficient than the 1st-order counterparts in some important applications; for example, fuzzy logic operations, as explained before [7].

Multi-layer feedforward networks [10], with at least three layers but neither shortcut nor closed loop, are quite common ANNs. Therefore, we choose this architecture as the target for extension of the traditional BP algorithm to the 2nd-order networks.

Specifically, the 2nd-order BP algorithm can be formulated as follows. We assume that the activation function for every neuron is the sigmoid function $\sigma(x) = \dfrac{1}{1+\exp(-x)}$, the network has $m$ inputs denoted as $\{\vec{x}_i \in R^m\}$ and $n$ outputs denoted as $\{\vec{y}_i \in R^n\}$. With $\vec{y}'$ as the final output of the feedforward propagation process, the loss function is expressed as:

$$E(\vec{w},\vec{b},\vec{c}) := \frac{1}{2}\sum_i (y_i' - y_i)^2 = \frac{1}{2}\sum_i (h_{\vec{w},\vec{b},\vec{c}}(\vec{x}_i) - y_i)^2, \quad (2)$$

which is proportional to the Euclidean distance between the vectors $\vec{y}'$ and $\vec{y}$. For a neuron at the $l^{th}$ layer with $p$ input variables, its output $o_k^l$ is an input to the $(l+1)^{th}$ layer, and can be computed by

$$\begin{aligned} o_j^{l+1} &= \sigma(\text{net}_j) \\ &= \sigma\left((\sum_{k=1}^p w_{kj\_r} o_k^l + b_{j\_r})(\sum_{k=1}^p w_{kj\_g} o_k^l + b_{j\_g}) + \sum_{k=1}^p w_{kj\_b}(o_k^l)^2 + c_j\right). \end{aligned} \quad (3)$$

The value of $\text{net}_j$ for activation by the neuron $j$ is a quadratic operation of $\vec{o}^l$, the outputs of $p$ neurons at the previous layer. For the neurons in the input layer, the inputs are our original data.

The goal of training the network is to adjust parameters for minimized $E$. The optimal parameters can be found using the gradient descent method starting from initial values. The derivative of the sigmoid function has a simple form:

$$\frac{d\sigma}{dz}(z) = \sigma(z)(1-\sigma(z)) \quad (4)$$

First of all, we separate the contribution of individual training sample:

$$\begin{cases} \dfrac{\partial E}{\partial \vec{w}} = \sum_i \dfrac{\partial E_i}{\partial \vec{w}} \\ \dfrac{\partial E}{\partial \vec{b}} = \sum_i \dfrac{\partial E_i}{\partial \vec{b}}, \\ \dfrac{\partial E}{\partial \vec{c}} = \sum_i \dfrac{\partial E_i}{\partial \vec{c}} \end{cases} \quad (5)$$

where the weights $w_{kj\_r}$, $w_{kj\_g}$, $b_{j\_r}$, $b_{j\_g}$, $c_j$ connect the $k^{th}$ neuron in the $(l-1)^{th}$ layer and the $j^{th}$ neuron in the $l^{th}$ layer. The partial derivatives of the loss function can be computed by the chain rule:

$$\begin{cases} \dfrac{\partial E_i}{\partial w_{kj\_r}} = \dfrac{\partial E_i}{\partial o_j^l} \dfrac{\partial o_j^l}{\partial \text{net}_j} \dfrac{\partial \text{net}_j}{\partial w_{kj\_r}} \\ \dfrac{\partial E_i}{\partial w_{kj\_g}} = \dfrac{\partial E_i}{\partial o_j^l} \dfrac{\partial o_j^l}{\partial \text{net}_j^l} \dfrac{\partial \text{net}_j^l}{\partial w_{kj\_g}} \\ \dfrac{\partial E_i}{w_{kj\_b}} = \dfrac{\partial E_i}{\partial o_j^l} \dfrac{\partial o_j^l}{\partial \text{net}_j^l} \dfrac{\partial \text{net}_j^l}{\partial w_{kj\_b}} \\ \dfrac{\partial E_i}{\partial b_{j\_r}} = \dfrac{\partial E_i}{\partial o_j^l} \dfrac{\partial o_j^l}{\partial \text{net}_j^l} \dfrac{\partial \text{net}_j^l}{\partial b_{j\_r}} \\ \dfrac{\partial E_i}{\partial b_{j\_g}} = \dfrac{\partial E_i}{\partial o_j^l} \dfrac{\partial o_j^l}{\partial \text{net}_j^l} \dfrac{\partial \text{net}_j^l}{\partial b_{j\_g}} \\ \dfrac{\partial E_i}{\partial c_j} = \dfrac{\partial E_i}{\partial o_j^l} \dfrac{\partial o_j^l}{\partial \text{net}_j^l} \end{cases} \quad (6)$$

If $j$ is the output node, that is, the $l^{th}$ layer is actually the output layer, then $o_j^l = h_{\vec{w},\vec{b},\vec{c}}(\vec{x}_i)$,

$$\frac{\partial E_i}{\partial o_j^l} = o_j^l - y_i. \quad (7)$$

If $j$ is not the output node, then the change of $o_j^l$ will affect all the neurons connected to it in the $(l+1)^{th}$ layer. Assuming $L$ is the set of indices for such neurons, the first factor becomes

$$\begin{aligned} \frac{\partial E_i}{\partial o_j^l} &= \sum_{v \in L} \left( \frac{\partial E_i}{\partial \text{net}_v^{l+1}} \frac{\partial \text{net}_v^{l+1}}{\partial o_j^l} \right) \\ &= \sum_{v \in L} \left( \frac{\partial E_i}{\partial \text{net}_v^{l+1}} \frac{\partial \text{net}_v^{l+1}}{\partial o_j^l} \right) \end{aligned} \quad (8)$$

For the 2nd order neurons, we have

$$\frac{\partial \text{net}_v^{l+1}}{\partial o_j^l} = w_{jv\_r} (\sum_{j=1}^n w_{jv\_g} o_j^l + b_{j\_g}) + w_{jv\_g} (\sum_{j=1}^n w_{jv\_r} o_j^l + b_{j\_r}) + 2\sum_{j=1}^n w_{jv\_b} o_j^l, \quad (9)$$

where $n$ is the number of the neurons in the $l^{th}$ layer. Furthermore, the second factor in Eq. (5) is:

$$\frac{\partial o_j^l}{\partial \text{net}_j^l} = \sigma(\text{net}_j^l)(1-\sigma(\text{net}_j^l)) = o_j^l(1-o_j^l), \quad (10)$$

Finally, the third factor is computed as

$$\begin{cases} \dfrac{\partial \text{net}_j^l}{\partial w_{kj\_r}} = o_k^{l-1}(\sum_{i=1}^{m} w_{ij\_g} o_i^l + b_{j\_r}) \\ \dfrac{\partial \text{net}_j^l}{\partial w_{kj\_g}} = o_k^{l-1}(\sum_{i=1}^{m} w_{ij\_r} o_i^l + b_{j\_g}) \\ \dfrac{\partial \text{net}_j^l}{\partial w_{kj\_b}} = (o_k^{l-1})^2 \\ \dfrac{\partial \text{net}_j^l}{\partial b_{j\_r}} = \sum_{i=1}^{m} w_{ij\_g} o_i^l + b_{j\_g} \\ \dfrac{\partial \text{net}_j^l}{\partial b_{j\_g}} = \sum_{i=1}^{m} w_{ij\_r} o_i^l + b_{j\_r} \end{cases} \quad (11)$$

where $m$ is the number of the neurons in the $l^{th}$ layer.

From Eqs. (5)-(11), all the partial derivatives can be computed to implement the gradient-based optimization.

Typically, the steepest descent algorithm in the following form is used:

$$\alpha = \alpha - \eta \cdot \dfrac{\partial E}{\partial \alpha}, \quad (12)$$

where $\alpha$ is a generic parameter, and $\eta$ is the step length. The optimization flowchart is in Fig. 2.

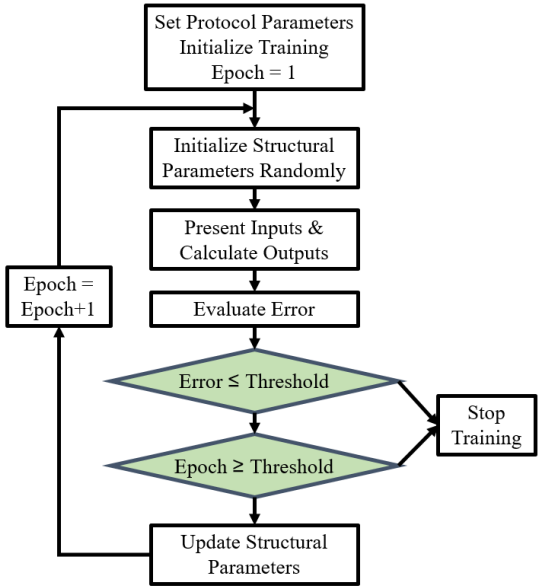

Fig. 2. Flowchart of the general BP algorithm to update structural parameters (weights including offsets).

## III. NUMERICAL RESULTS

To evaluate the power of 2nd-order networks, we started with testing the network architecture with two hidden layers. For illustration, the BP process can be more clearly illustrated with the double hidden layer network than with a single hidden layer network. The computational power of the double hidden layer network is believed sufficiently strong in this pilot study.

For visualization, the training datasets contained vectors of only two elements for the first two examples. The color map "cool" in MATLAB was used to show the functional values. The symbols "o" and "+" are labels for two classes respectively.

### A. "Telling-Two-Spirals-Apart" Problem

The "Telling-Two-Spirals-Apart" problem is a classical neural network benchmark test proposed by Wieland [11]. There are 194 instances on the two spiral arms, with 32points per turn plus an endpoint in the center for each spiral. The training points were generated with the following equations [12].

$$x^{2n-1} = r_n \sin \alpha_n + 0.5 = 1 - x^{2n} \quad (13)$$
$$x^{2n} = 0.5 - r_n \sin \alpha_n = 1 - x^{2n-1} \quad (14)$$
$$y^{2n-1} = r_n \cos \alpha_n + 0.5 = 1 - y^{2n} \quad (15)$$
$$y^{2n} = 0.5 - r_n \cos \alpha_n = 1 - y^{2n-1} \quad (16)$$
$$b^{2n-1} = 1 \quad (17)$$
$$b^{2n} = 0 \quad (18)$$
$$r_n = 0.4 \left( \dfrac{105 - n}{104} \right) \quad (19)$$
$$\alpha_n = \dfrac{\pi(n-1)}{16} \quad (20)$$

Lang and Witbrock reported that the standard backpropagation network with forward connections cannot classify such spirals. Then, they made a 2-5-5-5-1 network architecture with shortcut to solve this problem successfully. They trained the network with the BP algorithm and *QuickProp* [13].

In our experiment, however, we constructed the 2nd-order network of the 2-20-20-1 configuration without any shortcut, and trained the network with our generalized BP algorithm in a divide-and-conquer fashion (step by step, from the outer turn toward inner turn). The result is color-mapped in Fig. 3, showing the superior inherent power of the 2nd-order network.

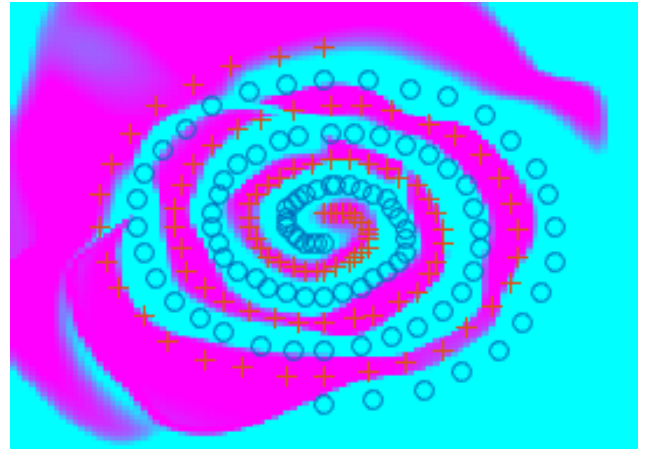

Fig. 3. Classification of two spirals with the 2nd-order network of the 2-20-20-1 configuration.

### B. Separating Concentric Rings

Another type of natural patterns is concentric rings. As a test, we generated four concentric rings, which were respectively assigned to two classes. There are 60 instances per ring. With randomly selected initial parameters, the 2nd-order network of a 2-3-2-1 configuration was trained, perfectly separating the two classes of instances in no more than 1,000 iterations, as shown in Fig. 4. In contrast, we also trained the 1$^{st}$-order networks. Even if we increased the complexity of the 1$^{st}$-order network to a 2-20-10-1 configuration, the circles could still not be classified well. It is seen in Fig. 5 that the purple ring is uneven with a defect zone at the bottom of the purple ring.

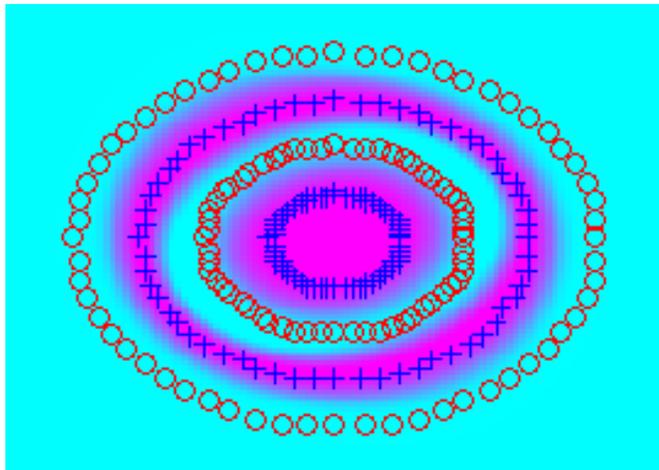

Fig. 4 Perfect classification of concentric rings by the 2$^{nd}$-order network of the 2-3-2-1 configuration.

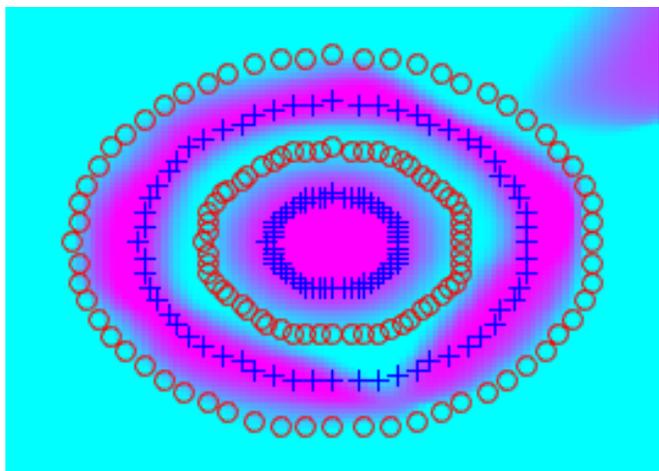

Fig. 5 Imperfect classification of concentric rings by the 1$^{st}$-order network of the 2-20-10-1 configuration.

*C. Sorting CT and MRI Images*

Computed tomography (CT) and magnetic resonance imaging (MRI) have many clinical applications. While both CT and MRI tomographically depict the structure and function of the patient, they have relative merits and drawbacks. CT enjoys high spatial resolution, little geometric distortion, fast speed, and cost-effectiveness. MRI is featured by high contrast resolution, rich functional information, pulse programming flexibility, and no radiation risk.

However, neither CT nor MRI is perfect. CT uses ionizing radiation that is potentially harmful to the patient if the radiation dose is not well controlled. Also, CT is not good at revealing subtle features of soft tissues. The recent development of photon-counting energy-discriminating CT promises to improve contrast resolution but this spectral CT technology is still under development. As far as MRI is concerned, the major problems are also multiple. First, a MRI scanner is significantly more expensive than a CT scanner. Also, MRI suffers from poor spatial resolution when its scanning speed is not too slow. Moreover, there is substantial geometric distortion in MRI images, due to various magnetic interferences. Recently, it has been reported that some contrast agents used for MRI may be potentially harmful, and have been stopped in Europe.

Our group proposed the conceptual design of a simultaneous CT-MRI scanner to integrate advantages of CT and MRI and correct their respective shortcomings for cardiac studies, cancer imaging, and other applications [14]. CT-MRI is the next step after PET-CT and PET-MRI, and eventually would lead to a PET-CT-MRI scanner, which we call "*Omnitomography*" for super-radiomics.

To perform image reconstruction for our intended CT-MRI scanner, we proposed an algorithmic framework emphasizing structural coupling between CT and MRI images [15]. In the same spirit, there are algorithms available in the literature for such estimation, which are either atlas based [16] or feature learning based [17]. In our joint CT-MRI reconstruction process, we utilize mutual information between CT and MRI data, among other priors, to improve CT-MRI reconstruction quality. However, our proposed utilization of mutual information has been limited to the case of a single joint CT-MRI scan. Given the great successes of deep learning, we are motivated to develop a dual mechanism for estimation of CT images from MRI images and vice versa in the machine learning framework.

Now, we are interested in collecting big CT data and big MRI data, which may or may not be matched. One convenient way to collect CT and MRI images automatically is to search, download, and archive CT and MRI images from Internet. As an initial step of this effort, we need to be able to sort CT and MRI images. Here we develop a preliminary 2$^{nd}$-order network for this purpose.

We downloaded 15 CT images and 15 MRI images from Internet and then extracted image patches of 28*28 pixels from these images. For classification, we constructed a 2$^{nd}$-order Convolutional Neural Network (CNN), which consists of a convolutional layer, a pooling layer and a full connected layer, as shown in Fig. 6. Then with random initial values, the network was trained using our general BP algorithm. After 10 iterations, the CNN can classify CT and MRI images successfully.

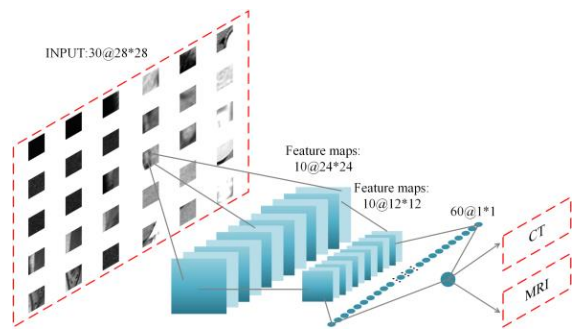

Fig. 6. Structure of the 2$^{nd}$-order CNN trained to sort CT and MRI images for construction of a big dataset of CT and MRI images.

Finally, we compared the performance of the 2$^{nd}$ CNN with that of the 1$^{st}$ CNN in the same architectures. One has 10 initial feature maps and 60 subsequent feature maps (10-60 configuration), and the other has 5 initial feature maps and 70 subsequent feature maps (5-70 configuration). The number of iterations was set to 10. We training the networks 20 times using randomly initialized values, and then recorded the number of images correctly labeled. With the 10-60 configuration, the average success rate was 72.17% for the 2$^{nd}$-order CNN, which is higher than 60.00% for the 1$^{st}$-order CNN. With the 5-70 configuration, the success rates were 76.50% for 63.83% for the 2$^{nd}$-order and 1$^{st}$-order networks respectively.

## IV. DISCUSSIONS AND CONCLUSION

Our successes in the "*Telling-Two-Spirals-Apart*" and other tasks have demonstrated that our generalized BP algorithm works well for training the 2$^{nd}$-order multilayer forward network. In principle, the BP principle can be adapted for any 2$^{nd}$-order networks. Nevertheless, it is still desirable to refine the BP algorithm for better efficiency and reliability, using techniques such as *Quikprop* [13] and learning rate adaption [18].

Based on our experience so far with 2$^{nd}$-order networks, it seems that the 2$^{nd}$-order network could be considerably simpler in the network topology, despite that individual neurons are more complicated. We believe that in certain applications, the overall "after-training" performance of the 2$^{nd}$-order network would be better than the 1$^{st}$-order counterpart in terms of accuracy and speed for a given computational complexity; for example, in the case of separating concentric rings the 1$^{st}$-order network is inherently handicapped. Recurrent neural networks [19] consisting of 1$^{st}$ neurons were demonstrated to perform well for approximation of nonlinear functions. RNNs with 2$^{nd}$-order neurons should be more flexible.

As a biomedical imaging related example, we have shown that a 2$^{nd}$-order network can sort CT and MRI images from Internet. Next, we will enhance this network to not only differentiate between CT and MRI images but also reject those images that are of neither CT- nor MRI-type. This will lead to a big dataset consisting of numerous CT and MRI images from Internet. We believe that these images can be registered to some standard CT and MRI atlases such as visual human CT and MRI volumes. Then, this huge atlas should be able to help improve prediction from CT images to MRI images, and vice versa, and facilitate joint CT-MRI reconstruction as well.

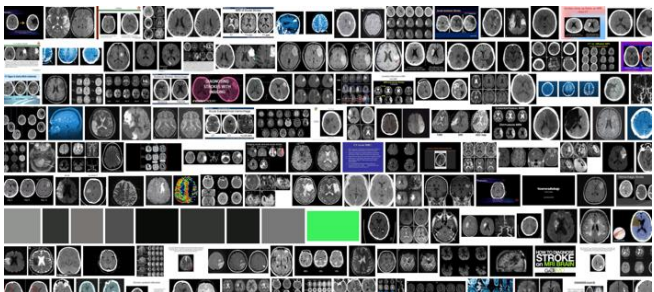
Fig. 7. Idea to form a CT/MRI ImageNet via a 2$^{nd}$-order neural network.

In conclusion, the general BP algorithm has been formulated and tested for the 2$^{nd}$-order network in this pilot study. Numerical examples show the BP algorithm works well in some biomedical engineering relevant cases. Much more efforts are needed to refine the optimization method and design the 2$^{nd}$-order networks, and find killer applications in the real world.